# On the Effects of Different Types of Label Noise in Multi-Label Remote Sensing Image Classification

Tom Burgert, *Member, IEEE*, Mahdyar Ravanbakhsh, *Member, IEEE*, and Begüm Demir, *Senior Member, IEEE*

*Abstract*—The development of accurate methods for multi-label scene classification (MLC) of remote sensing (RS) images is one of the most important research topics in RS. To address MLC problems, the use of deep neural networks that require a high number of reliable training images annotated by multiple land-cover class labels (multi-labels) has been found popular in RS. However, collecting such annotations is time consuming and costly. A common procedure to obtain annotations at zero labeling cost is to rely on thematic products or crowdsourced labels. As a drawback, these procedures come with the risk of label noise that can distort the learning process of the MLC algorithms. In the literature, most label noise robust methods are designed for single-label classification (SLC) problems in computer vision (CV), where each image is annotated by a single label. Unlike SLC, label noise in MLC can be associated with: 1) subtractive label noise (a land cover class label is not assigned to an image while that class is present in the image); 2) additive label noise (a land cover class label is assigned to an image, although that class is not present in the given image); and 3) mixed label noise (a combination of both). In this article, we investigate three different noise robust CV SLC methods (self-adaptive training (SAT), early-learning regularization, and joint co-regularized training) and adapt them to be robust for multi-label noise scenarios in RS. During experiments, we study the effects of different types of multi-label noise and evaluate the adapted methods rigorously. To this end, we also introduce a synthetic multi-label noise injection strategy that is more adequate to simulate operational scenarios compared to the uniform label noise injection strategy, in which the labels of absent and present classes are flipped at uniform probability. Further, we study the relevance of different evaluation metrics in MLC problems under noisy multi-labels. On the basis of the theoretical and experimental analyses, some guidelines for a proper design of label noise robust MLC methods are derived.

## I. INTRODUCTION

RISING interest in earth observation has led to great technological advances, increasing the availability of remote sensing (RS) image archives in recent years. The development of multi-label scene classification (MLC) methods belongs to one of the prominent research directions in the RS community. Recent works in MLC follow up with the advances in computer vision (CV) and employ state-of-the-art deep learning approaches [1], [2], [3]. However, deep neural networks generally require large amounts of training data. Acquiring manual annotations to carry out supervised classification tasks remains an expensive procedure. Therefore, more and more annotation processes involve crowdsourcing procedures or rely on exploiting publicly available thematic products like Corine Land Cover (CLC) map [4], GLC2000 [5] or GlobCover [6]. While cheaper to obtain, these approaches come with the risk of label noise. For example, in the case of crowdsourced data, non-expert annotators might mislabel images due to a lack of knowledge, while making use of thematic products can induce noise due to imprecise mappings or outdated data.

Initial works in RS label noise robust learning have been proposed for single-label learning problems [7], [8], [9]. However, according to our knowledge, there are only two methods presented in the context of multi-label noise robust classification problems in RS. In detail, Hua et al. [10] propose to regularize the network predictions by a label correlation matrix derived from distances between corresponding word embeddings of class labels, while Aksoy et al. [11] introduce a collaborative learning framework to discard noisy-labeled examples during training. In [11], by loss function design, the two neural networks are forced to produce distinct features via a discrepancy module while at the same time ensuring similar predictions through a consistency loss. When training images are associated with multiple labels, noise can arise in two different ways: 1) noise related to missing labels: i.e., a land cover class is present in the image but is not annotated as a label, in the following referred to as *subtractive noise* or 2) noise related to false labels: i.e., a land cover class is not present in the image but is annotated as a label, in the following referred to as *additive noise* (see Fig. 1). None of the RS works study the effect of these types of label noise separately. The robustness of methods in the RS literature is merely evaluated by injecting uniform label noise, in which each class label of a given image has the same probability of being flipped from absent to present or vice versa. Since for multi-label training sets, the majority of classes are not present in an image (as an example, see classes vs. average classes per image in Table II), injecting uniform noise leads to a skewed distribution of overly induced *additive noise* and little induced *subtractive noise*. Any result can be assumed to be dominated by the effects of *additive noise*.

Unlike RS, label noise problems have been extensively studied in the CV community, but due to the differences in

Manuscript received 13 July 2022; revised 9 November 2022; accepted 21 November 2022. Date of publication 1 December 2022; date of current version 19 December 2022. This work was supported in part by the German Federal Ministry of Education and Research through the TreeSatAI Project under Grant 01IS20014A and in part by the European Research Council (ERC) through the ERC-2017-STG BigEarth Project under Grant 759764. *(Corresponding author: Begüm Demir.)*
The authors are with the Faculty of Electrical Engineering and Computer Science, Technische Universität Berlin, 10623 Berlin, Germany (e-mail: t.burgert@tu-berlin.de; ravanbakhsh@tu-berlin.de; demir@tu-berlin.de).
Digital Object Identifier 10.1109/TGRS.2022.3226371

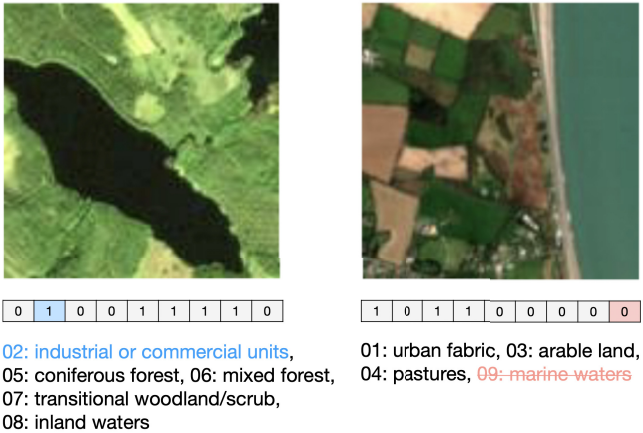

Fig. 1. Example of multi-label noise in an RS image. The binary vector below each image shows the label set of that image. Each element of the vector indicates the presence or absence of a class. (Left) *Additive noise*. The label vector incorrectly indicates "industrial or commercial units" as a present class. (Right) *Subtractive noise*. The label vector incorrectly omits "marine waters" as a present class.

image structure and availability of labels, mainly in single label classification (SLC) problems (see Section II for a comprehensive summary). In this article, we aim to extend the knowledge gathered about single-label noise robust research to MLC in RS. To this end, we discuss the potentials and limitations of noise robust SLC methods in RS MLC. It is worth noting that by design, not every noise robust method developed for SLC is applicable to RS MLC problems. Based on their suitability, we investigate the three SLC methods: 1) self-adaptive training (SAT) [12]; 2) early-learning regularization [13]; and 3) joint co-regularized training [14]. Particularly, we adapt them to handle multi-label noise and evaluate them under the different types of label noise. Further, we propose a novel noise injection strategy that allows to simulate noisy annotation processes more adequately. In detail, it enables the study of the effects of *additive* and *subtractive noise* separately and considers an equal distribution of both types when studying them jointly (*mixed noise*).

In summary, our contributions are the following.

1) To the best of our knowledge, we are the first to present and discuss how to adapt noise robust methods from SLC to MLC problems. In particular, we present the adaptations of: i) SAT [12]; ii) early-learning regularization [13]; and iii) joint co-regularized training [14] to handle multi-label noise.
2) We present a novel noise injection strategy that overcomes the limitations of existing uniform noise injection approaches and thus leads to a realistic evaluation of the robustness of MLC methods to different types of label noise for operational MLC problems in RS.
3) We provide a critical performance analysis of: i) the effects of different noise types in MLC problems; ii) the relevance and importance of different evaluation metrics; and iii) the adapted methods under different label noise types on three multi-label RS datasets. Then, we provide guidelines for the proper design and development of multi-label noise robust methods.

The rest of the article is organized as follows. In Section II, we survey the related noise robust methods for both SLC and MLC in CV. In Section III, the three noise robust methods adapted from SLC to MLC are introduced in detail. Section IV describes the considered datasets and the experimental setup, while Section V represents the experimental results. Finally, in Section VI, the conclusions of the work are drawn.

## II. Related Work

In this section, we review related noise robust methods for both SLC and MLC developed in the CV community. In SLC, the label of an image is considered to be noisy when its associated class does not coincide with its ground truth class. In MLC, noise can arise in two different ways. The label set can either indicate the presence of a class that is actually absent in the image (*additive noise*) or the absence of a class that is actually present in the image (*subtractive noise*). In general, label noise research in CV is mainly focused on SLC, leading to strong methodological papers and unified evaluation procedures under common benchmark datasets and noise injection strategies. The less investigated research field of multi-label noise is still missing such standardization.

### A. Noise Robust Methods in Single-Label Image Classification

There are two main groups of approaches aiming to learn from noisy-labeled data for SLC problems. Methods belonging to the first of the two groups assume the existence of a clean and trustworthy subset of data. There are various techniques to enable noise robust learning by exploiting this subset. Hendrycks et al. [15] use the clean data to estimate a noise transition matrix to improve the predictions of a classifier trained on the noisy-labeled data. Lee et al. [16] build class prototypes in feature space from the clean subset of data. During training, these prototypes are then compared to the images from the noisy-labeled training set, resulting in different similarity-based attention weights for the images. In contrast, Li et al. [17] use the clean data to train an auxiliary model leveraged by a knowledge graph that produces additional soft labels for the noisy-labeled training set, contributing to the standard loss function of the original labels.

The second group of approaches aims to achieve more generalization by not utilizing any clean data and learning directly from the noisy-labeled data. Some approaches in this group propose specific design choices for training that are inherently noise robust. This is mainly achieved by specialized architectures [18], [19] and noise robust loss functions [20], [21], but can also include semi-supervised data augmentation techniques. Zhang et al. [22], for example, generate additional data by convex combinations of pairs of images and their labels, leading to more noise robustness during training. Yet, the majority of approaches belonging to the second group perform some sort of auxiliary task for handling noise during training. Such tasks usually extend the classic training procedure of a neural network by a specific method or subroutine that is designed to detect and handle label noise. These routines either perform a process of reweighting or discarding noisy-labeled examples (i.e., images) [14], [23], [24], [25], [26], [27], [28], [29] or, alternatively, incorporate noise specific information into a standard loss function [12], [13], [17], [30], [31]. The information for noise handling is derived by exploiting prediction values [12], [13], [24], [27], feature representations [17], [23], [30], or additional information from an ensemble of networks. Ensembles of networks may be used in the form of collaborative networks [14], [25], [26], [28], student–teacher networks [17], [29], [31], or as a self-ensemble network [24], [31].

In particular, Guo et al. [23] design a training curriculum based on complexity clusters. The authors show that most examples with noisy labels get assigned to clusters with higher complexities and, hence, only influence the training process at later stages, at which the model has learned the dominant patterns already. Han et al. [30] build multiple prototypes for each class that are used to produce corrected pseudo labels based on a similarity measure in feature space during the training process. The loss function of the neural network is mutually influenced by the observed and corrected labels. Following the strategy of sample selection, Malach and Shalev-Shwartz [25] maintain two networks that are being updated only if their predictions disagree. The strategy assumes that correctly labeled hard examples are more likely to produce ambiguous predictions than mislabeled easy examples. Inspired by this work, Wei et al. [14] train two networks with small loss instances only, which are derived from a joint loss to ensure the agreement of both networks. Addressing noise robustness via loss function design, Huang et al. [12] employ a label refurbishment loss that makes use of the exponential moving average of predictions to progressively correct wrong labels. Further, Liu et al. [13] exploit prediction values from the early learning phase to impose a regularization term to the standard loss to neutralize the gradient of examples with false labels. Li et al. [31] propose to incorporate information on mislabeled examples into the loss function within a meta-learning step. During this phase, multiple mini-batches of synthetic label noise are generated by a random neighbor label transfer. Each synthetic mini-batch updates a copy of the current model enforcing consistent predictions with a self-ensemble teacher model and all meta-updated models at this stage.

For a more in-depth overview of noise robust methods in SLC, we refer the reader to the survey article of Song et al. [32].

### B. Noise Robust Methods in Multi-label Image Classification

While research on noise robust methods for SLC is well established in the CV community, with consistent definitions of label noise and benchmark datasets, most research conducted in the multi-label case appears to be more heterogeneous. Some research fields treat multi-label noise partially without explicitly mentioning it. For example, the research field of multi-label learning with missing labels (MLML) can be interpreted as a problem inducing *subtractive noise* for MLC. In MLML, labels can be either annotated as present or absent or not annotated at all (missing). A common approach to deal with the missing labels is to consider them as absent [33]. This procedure induces *subtractive noise* in the form of present classes that are annotated as absent. Bucak et al. [34] solve the problem of MLML using group lasso techniques and a ranking loss applied to a support vector machine classifier, while Jain et al. [35] introduce a propensity-scored loss function that is used to train a tree classifier. Further, Durand et al. [33] propose a version of the binary cross entropy (BCE) loss function that adapts itself to the proportion of known labels per sample. The loss function is used to train a state-of-the-art convolutional network whose output is concatenated with a graph neural network to model the correlation between classes.

On the other hand, the research field of partial multi-label learning (PML), which aims at selecting the correct labels out of a set of candidate labels, can be interpreted as the counterpart of MLML. This task induces a form of *additive noise*. Incorrect candidate labels are equivalent to noise in form of absent classes that are annotated as present. Xie and Huang [36] rank the candidate labels such that true labels are ranked higher than noisy labels. This is achieved by making use of a confidence-weighted ranking loss enhanced by label correlations and feature prototypes.

It exists a small body of literature that deals with training sets that are naturally noisy, comprising both types of label noise [37], [38], [39]. All three works follow a similar approach of training a label-cleaning network based on a small clean subset of data that subsequently supervises the training process of a second classification network with noisy-labeled data. While the experiments are conducted only on training sets whose labels are inherently noisy, a comprehensive study of the behavior under different noise rates is missing. Moreover, an additional limitation to these approaches is the restriction to datasets with clean subsets available. Considering both types of noise, Kumar et al. [40] provide theoretical proof that the mean absolute error is noise robust in multi-label scenarios under the assumption that ground truth present and absent labels are corrupted with the same probability. However, this assumption over-represents noise in the form of present classes that are actually absent since the label matrices are mostly sparse with a distribution heavily skewed toward present labels. Still, the only work that studies both types of noise at the same time and evaluates it under different noise rates considers uniform noise [41]. Zhao and Gomes [41] utilize an encoder–decoder architecture, where the decoder part consists of an attention-based graph neural network, with each node corresponding to a label represented by an embedded vector. The learning signal is composed of a BCE loss on the output of the graph neural network and a regularization term for the learnable label embeddings.

## III. Adaptation of SLC to MLC Methods

Toward noise robust methods in MLC, the well-established landscape of research conducted in CV considering single-label noise can reveal promising design choices for dealing with multi-label noise. In this section, we discuss the potential of adaptations of noise robust methods from SLC to MLC. In particular, we present the theoretical background of three methods that we adapt to handle RS images annotated with multi-labels.

### A. Problem Definition

Let $\mathcal{D} = \{(\mathbf{x}_1, \mathbf{y}_1), \ldots, (\mathbf{x}_N, \mathbf{y}_N)\}$ be a multi-label training set with $C$ classes that consists of $N$ tuples of images $\mathbf{x}_i$ associated with label sets $\mathbf{y}_i$. Each label set is defined as a binary vector $\mathbf{y}_i \in \{0, 1\}^C$, where each element $y_{i,c}$ is indicating the presence or absence of a class $c$ in the image $\mathbf{x}_i$. Under multi-label noise, *additive noise* refers to a label set entry $y_{i,c} = 1$ for a class that is absent in the image, *subtractive noise* refers to a label set entry $y_{i,c} = 0$ for a class that is present in the image.

### B. From Single-Label to Multi-label Noise Robust Methods

In general, not every noise robust approach developed for single-label scenarios is feasible to be adapted to handle multiple labels. For example, hybrid approaches that use data-augmentation techniques to pseudo-label the noisy-labeled data [42], [43] suffer from the limitations multiple labels impose on augmentation techniques. Any augmentation strategy that omits areas of the original image, including random crop, cut-out, scaling, translation, as well as most rotations, could remove classes that are located close to the borders of the image. It remains an open question whether these augmentation techniques can still supply enough information to noise detection strategies when constrained by multi-label requirements. On the other hand, any strategy that includes class-based decisions that elaborate information from embedded features, such as prototyping [30], clustering [23], or synthetic class label transfer [31], is impracticable. Feature representations of examples with multiple labels decode abstract class information of different subsets of classes at once—a disjoint grouping by individual classes is not possible. Conversely, treating each combination of present classes as an independent class grows exponentially with the number of classes, already exceeding 1000 combinations for datasets with just ten classes. Besides the neglect of semantic correlation between combinations with intersecting classes, this approach leaves many combinations underrepresented and impedes a reasonable application in MLC.

However, any approach that elaborates information derived from prediction values or ensembles can be adapted to handle multiple labels easily. As a step toward noise robust methods in MLC in RS, we choose three diverse approaches from SLC in CV, adapt them, and evaluate their robustness for multi-label noise. In the following, the theory of these approaches, together with the choice of adaptation, is described briefly. A supportive tabular comparison can be found in Table I. For

TABLE I

Categorization of the Considered Noise Robust Methods and Their Adaptation Requirements to Handle Multi-Label Noise. (B): Basic Adaptation, Replace Cross Entropy Loss by Binary Cross Entropy Loss and Softmax Layer by Sigmoid Layer. (R): Regularization Adaptation, Resolve Regularization Term From Example-wise to Label-wise. (S): Selection Adaptation, Select Individual Label Losses Instead of Entire Example Losses

| Method | Group | Subgroup | Adaptation |
|---|---|---|---|
| SAT [12] | Loss Adjustment | Label Refurbishing | B |
| ELR [13] | Sample Selection | Ensemble-Learning | B, R |
| JoCoR [14] | Loss Adjustment | Regularization | B, S |

SLC problems, the associated label set $\mathbf{y}_i$ for single-labeled image $\mathbf{x}_i$ remains notated as a binary vector of size $C$ in which the multi-hot encoding of MLC is replaced by a one-hot encoding. Further, $f$ is defined to be a neural network parameterized by $\theta \in \Theta$ that maps images $\mathbf{x}_i$ to the label probability distribution $\mathbf{p}_i = f(\mathbf{x}_i, \theta)$ with entries $p_{i,c}$ indicating the predicted probability for class $c$. All three adaptations to MLC have in common that any cross entropy loss (1) is replaced by a BCE loss (2), and any softmax function is replaced by a sigmoid function

$$L_{\text{CE}}(\Theta) = -\frac{1}{n} \sum_{i=1}^{n} \sum_{c=1}^{C} y_{i,c} \log p_{i,c} \quad (1)$$

$$L_{\text{BCE}}(\Theta) = -\frac{1}{n} \sum_{i=1}^{n} \sum_{c=1}^{C} y_{i,c} \log p_{i,c} + (1 - y_{i,c}) \log(1 - p_{i,c}). \quad (2)$$

### C. Self-Adaptive Training

SAT is a method proposed by Huang et al. [12] that is designed to alleviate the overfitting issue of empirical risk minimization to improve the generalization of deep networks under label corruption. It can be considered as a label refurbishing strategy that enhances the loss function by making use of an exponential moving average of prediction values as targets. After a short warm-up phase that consists of training on the original labels, training targets are computed as a convex combination of labels and predictions. Concretely, given a data pair $(\mathbf{x}_i, \mathbf{y}_i)$ at a time step $t$, annotated as $(\mathbf{x}_i^{(t)}, \mathbf{y}_i^{(t)})$, and a prediction $\mathbf{p}_i^{(t)} = f(\mathbf{x}_i^{(t)}, \theta)$ with $t > E_s$, where $E_s$ is the number of epochs for a warm start, the corresponding updated target at time step $t+1$ used in the loss function is defined as follows:

$$\mathbf{y}_i^{(t+1)} = \alpha \cdot \mathbf{y}_i^{(t)} + (1 - \alpha) \cdot \mathbf{p}_i^{(t)} \quad (3)$$

where the momentum term $\alpha$ controls the weight on the model predictions. It is usually set around 0.9. $E_s$ is best set to correspond to the epoch at which the standard training procedure starts to overfit the corrupted labels. Generally, it applies that the higher the noise rate, the lower the optimal value for $E_s$. A disadvantage of SAT remains in the application of extremely small models that potentially underfit the data.

Standard empirical risk minimization outperforms the SAT method for models that are 10x smaller than ResNet-18 [44]. On the other hand, an adaptation to multi-labels is straightforward. The cross entropy loss (1) is exchanged by a BCE loss (2), while the softmax function to generate the prediction value is replaced by a sigmoid function.

### D. Early-Learning Regularization

Liu et al. [13] introduce a regularization term that pushes the model toward targets produced by semi-supervised learning techniques on the model outputs. In particular, it boosts the gradients of examples with clean labels and neutralizes the gradients of the examples with false labels, implicitly preventing memorization of the false labels. The choice of the regularization term is based on the assumption that the true class tends to be dominant in the earlier predictions for all examples. Let $\mathbf{p}_i$ be the prediction and $\mathbf{y}_i$ be the label for the $i$th example with the corresponding gradient of the cross entropy as follows:

$$\nabla L_{\text{CE}}(\Theta) = \frac{1}{n} \sum_{i=1}^{n} \nabla_{\Theta}(\mathbf{p}_i - \mathbf{y}_i). \quad (4)$$

For clean labels, the cross entropy term $\mathbf{p}_i - \mathbf{y}_i$ tends to vanish after the early-learning phase because $\mathbf{p}_i$ converges toward $\mathbf{y}_i$, eventually allowing the wrong labels, that are not yet memorized, to dominate the gradient. To counteract this effect, the following term is added to the cross entropy loss as follows:

$$L_{\text{ELR}}(\Theta) = \frac{\lambda}{n} \sum_{i=1}^{n} \log(1 - \langle \mathbf{p}_i, \mathbf{y}_i \rangle). \quad (5)$$

The authors show that the regularization term grows proportionally with increasingly learned clean examples at earlier stages, stagnating just before overfitting on noisy-labeled examples. After the early-learning phase, the term continues to amplify the gradient of clean labels while neutralizing the gradient of false labels. For detailed theoretical proofs, we refer the readers to the original paper [13]. By adapting the cross entropy loss (1) to the BCE loss (2), the regularization term needs to be adapted from example-wise to label-wise regularization. This can be achieved by resolving the dot product in the early-layer regularization (ELR) term, and simply adding

$$L_{\text{ELR-ML}}(\Theta) = \frac{\lambda}{n} \sum_{i=1}^{n} \sum_{c=1}^{C} \log(1 - p_{i,c} \cdot y_{i,c}) \quad (6)$$

for each binary class prediction $c$ independently.

### E. Joint Co-Regularized Training

Solving the task of learning under label noise via sample selection, Wei et al. [14] derive noise-specific information from an ensemble of networks by joint co-regularized training (JoCoR). In particular, the method maintains two pseudo-siamese networks $f$, $g$ that possess different learnable parameters $\theta_1$, $\theta_2 \in \Theta$, but are updated simultaneously by a joint loss. The loss is composed of three different components.

An individual cross entropy loss (1) is applied to each of the networks predictions $f(\mathbf{x}_i, \theta_1)$ and $g(\mathbf{x}_i, \theta_2)$. Additionally, a contrastive loss ensures co-regularization between the predictions of the two networks to maximize their agreement. The authors adopt the symmetric Kullback–Leibler (KL) divergence as the contrastive loss by

$$\begin{aligned} L_{\text{con}}(\Theta) = & D_{KL}[f(\mathbf{x}_i, \theta_1) \| g(\mathbf{x}_i, \theta_2)] \\ & + D_{KL}[g(\mathbf{x}_i, \theta_2) \| f(\mathbf{x}_i, \theta_1)]. \end{aligned} \quad (7)$$

Based on a weighted sum of the three losses, the small loss examples are selected for updating the networks. The design choice follows the assumption that the two networks would agree on most clean examples and are unlikely to agree on examples with incorrect labels. A potential side effect of the strategy could result in the rejection of difficult (but informative) examples for which both networks disagree due to high uncertainty. To enable multiple predictions at once, we exchange all cross entropy losses (1) by a BCE loss (2) as well as softmax layers by sigmoid layers. Further, the selection of small loss examples is replaced by a selection of small loss labels.

## IV. DATASET DESCRIPTION AND EXPERIMENTAL DESIGN

In the experiments, we used three RS multi-label datasets, namely: 1) UCMerced Land Use Dataset [45] denoted as UCMerced; 2) the initial version of the TreeSatAI dataset presented in [46] denoted as TreeSatAI; and 3) BigEarthNet19-Ireland dataset [47] denoted as BEN19-Ireland. Fig. 2 shows example images of the dataset. Further, a comparison of the characteristics of the three datasets can be found in Table II. In the following, the datasets are introduced briefly.

### A. Datasets

*1) UCMerced Dataset:* The multi-label version of the UCMerced dataset [45] consists of 2100 RGB images that were extracted from the USGS National Map Urban Area Imagery collection for various urban areas around the United States of America. We divided the dataset into 1890 train and 210 test images. The dataset is composed of 17 diverse classes ranging from vehicle classes like cars, boats, and airplanes to natural land cover classes like water and trees as well as buildings or pavements. Each image is a section of $256 \times 256$ pixels with a spatial resolution of 0.3 m. On average, each image is annotated with more than three present classes. Each class is at least $100\times$ present in the dataset. The dataset was manually labeled and guarantees clean annotations. There also exists a single-label version of the dataset [48].

*2) TreeSatAI Dataset:* In our experiments, we used the initial version of the TreeSatAI dataset that consists of 29 180 aerial images acquired from 2012 to 2020 across the German federal state of Lower Saxony [46]. Each aerial image is a section of $300 \times 300$ pixels, including RGB and near-infrared bands with a pixel resolution of 0.2 m. We divided the dataset into 19 995 train and 9185 test images. The annotations of the dataset are associated with 19 different tree species that were collected by experts through field surveys or photograph interpretations. While the dataset can be considered to be balanced (see Fig. 3), its label distribution resembles a SLC dataset. On average, there are only 1.5 classes annotated per image.

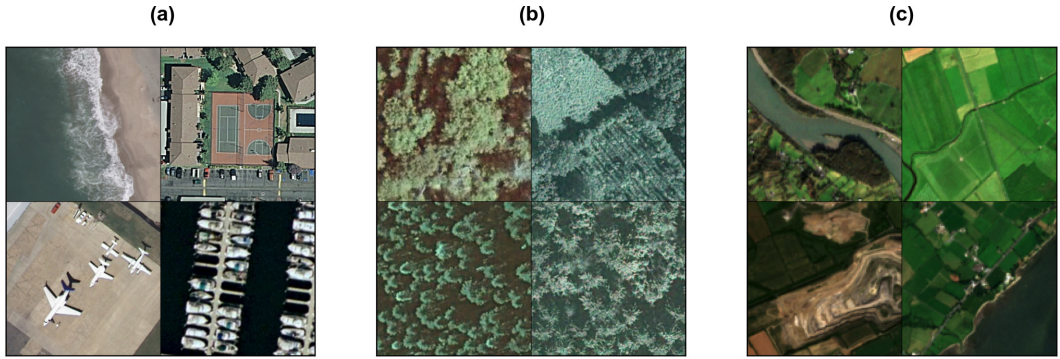

Fig. 2. Example images of (a) UCMerced dataset; (b) TreeSatAI dataset; and (c) BEN19-Ireland dataset.

TABLE II
CHARACTERISTICS OF THE MULTI-LABEL SCENE CLASSIFICATION DATASETS USED IN OUR EXPERIMENTS. L = LOW, M = MEDIUM, H = HIGH. * SUBJECTIVE RATINGS BASED ON CLASS DISTRIBUTION (SEE FIG. 3)

| Dataset | $|\mathcal{D}|$ | #Classes | Avg. #Classes per Image | #Bands | Image Sizes (Spatial Resolution) | Imbalance Factor* |
| --- | --- | --- | --- | --- | --- | --- |
| UCMerced [48] | 1,890 | 17 | 3.3 | 3 | 256x256 (0.3m) | M |
| TreeSatAI [46] | 19,995 | 19 | 1.5 | 4 | 300x300 (0.3m) | L |
| BEN19-Ireland [47] | 25,256 | 18 | 2.2 | 10 | 120x120 (10m), 60x60 (20m) | H |

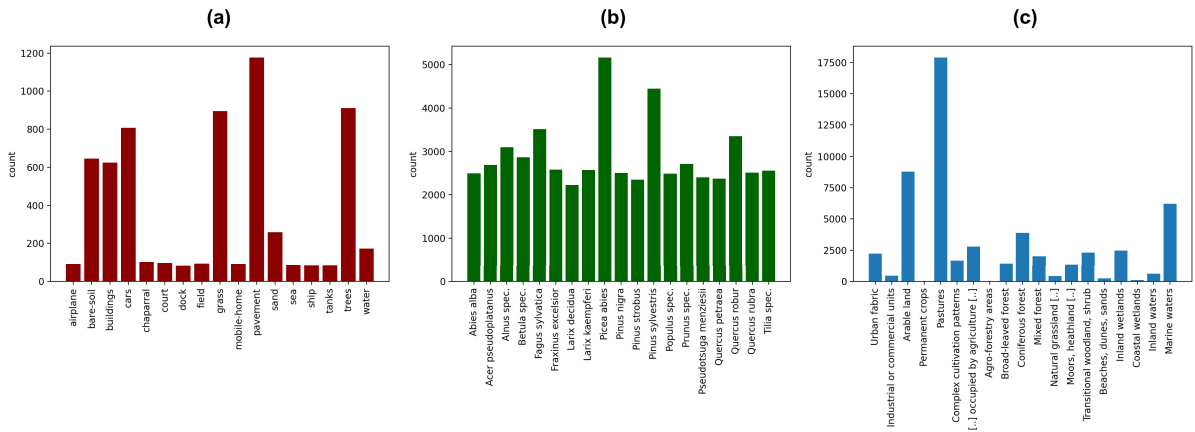

Fig. 3. Class distributions of (a) UCMerced dataset; (b) TreeSatAI dataset; and (c) BEN19-Ireland dataset.

*3) BEN19-Ireland Dataset:* BigEarthNet [47] is a multi-label dataset based on Sentinel-2 multispectral images, whose class annotations were automatically inferred by the 2018 CLC Map inventory. The originally 13-band images were captured from ten different countries in Europe. Considering the size of the dataset, the experiments are rolled out on the Ireland subset of BigEarthNet only. Adopting the land-cover class nomenclature proposed in [49], the Ireland subset consists of 25 256 train and 12 013 test images annotated with 18 land cover classes that include, e.g., different types of forests, buildings, or water. In particular, all classes cover consistent areas of the surface of the earth and, in contrast to the UCMerced dataset, exclude individual objects. In the following, the subset is referred to as BEN19-Ireland. During experiments, ten bands were considered that comprise a pixel resolution of 10 or 20 m at an image size of up to $120 \times 120$ pixels. While in average 2.2 classes are annotated per image, the class distribution of annotated present classes is heavily skewed, as it is shown in Fig. 3. Multiple classes are annotated for less than 1% of the images. Due to the use of the CLC product, this dataset may contain small portions of label noise naturally by the way it is constructed.

### B. Experimental Setup

For all of our experiments, including the adapted noise robust methods, we used the same training set-up to ensure a fair comparison. We chose a ResNet-18 [44] architecture as a backbone for all methods. The baseline model comprises the backbone architecture with a BCE loss. For the sake of simplicity, we denote the baseline model as BCE hereafter. Besides using pretrained weights on ImageNet [50], we did not apply any additional changes to the basic architecture provided by PyTorch. Further, we did not apply any data augmentation to the data. We trained the networks with an AdamW [51] optimizer with a weight decay of $1 \times 10^{-4}$ over 30 epochs, with an initial learning rate of $1 \times 10^{-3}$ that is reduced by a factor five after 20 epochs and a batch size of 64. Each score that is being presented reflects the average evaluation performance on the test sets of three independent training runs with different random seeds on different samples of noisy label sets under the same noise parameters.

## V. EXPERIMENTAL RESULTS

### A. Noise Injection Strategy for Evaluating Robustness

In our proposed multi-label noise injection strategy to evaluate the robustness of the adapted methods, the synthetically injected noise rates are orientated at the absolute number of present labels to model a noisy annotation process more realistically. In particular, for *additive* and *subtractive noise*, the percentage of entries from the label sets $\mathbf{y}_i$ that are being flipped relate to the absolute number of present labels in the label matrix $\mathcal{Y}$. Further, the noise is injected class-wise, preserving the original class distribution of present labels and preventing the quantification of artifacts arising from different class distributions. The injection procedure ensures that the effects of both noise types become more comparable to each other since each of them injects noise to the same amount of labels given a certain noise rate. Following the proposed strategy, an example of inducing 20% *mixed noise* to a training set can be seen in Fig. 4. In contrast to the existing literature, our noise injection strategy enables the study of the effects of *additive* and *subtractive noise* both separately and at the same time (*mixed noise*) with equal contribution of individual noise types regarding the absolute flips of labels.

### B. Effects of Different Noise Types

To better understand the effects of different noise types, we evaluate our baseline model for three different scenarios: purely *additive noise*, purely *subtractive noise*, and *mixed noise*, comprising both types of noise at the same time. We evaluate the model performance for synthetic noise rates between 10% and 70% with the mean average precision (mAP) metric. In particular, we compute scores for the metric globally by considering each element of a label set as an individual label (micro) and as an unweighted mean of the class-wise scores (macro). Fig. 5 shows the results evaluated for the three multi-label RS datasets for the baseline model. This plot indicates a variation in the effects of different noise types on the model performance. While the injection of *additive noise* has little impact on the model performance (consistent over all datasets), the injection of *subtractive noise* can lead to a greater decrease in performance. In particular, this is the case for the TreeSatAI dataset, which has the lowest number of average present labels (1.5). Micro and macro averaging both reveal such an effect. The highly imbalanced BEN19-Ireland dataset follows this trend in the macro scores only. In contrast, the UCMerced dataset (3.3 average present labels) does not reveal such a clear effect. Here, for most noise rates, *additive* and *subtractive noise* impose similar difficulty to the model. Only at a noise rate of 70% *subtractive noise* seems to have a greater negative effect. Taking into consideration that TreeSatAI incorporates the lowest fraction of present labels, the results suggest that *subtractive noise* can be more challenging to multi-label learning than *additive noise* if present labels are sparse. The hypothesis is further emphasized by the macro scores of BEN19-Ireland, which comprises a few underrepresented classes with a very small fraction of present labels. Since micro scores are dominated by classes with more present labels, the effect of *subtractive noise* impacting these underrepresented classes is only visible when observing macro scores. Generally, it can be noticed that in contrast to an individual type of noise, the mutual occurrence of *additive* and *subtractive noise* (*mixed noise*) always causes a significant high reduction in performance regardless of the dataset or metric. The above-described tendencies also hold for the adapted noise robust methods (see results in Tables III–V).

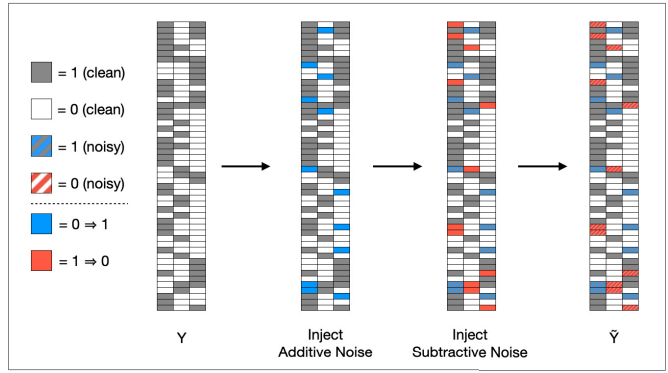

Fig. 4. Example of inducing 20% *mixed noise* (20% *additive noise* and 20% *subtractive noise*). $\mathcal{Y}$ is the clean label matrix each row representing label sets $y_i$, each column representing a class $c$. $\tilde{\mathcal{Y}}$ represents the noisified label matrix. *Additive noise* is depicted in blue changing an entry in the label set $y_i$ from 0 to 1, while *subtractive noise* is depicted in red changing an entry in the label set $y_i$ from 1 to 0.

### C. Relevance of mAP Metric

MLC competitions in CV, such as the PascalVOC challenge [52], are typically carried out on the mAP metric. Even though mAP and F1-score often correlate, there exist situations when this is not the case. By enabling the prediction of multiple classes at the same time, the predicted probabilities from individual classes are decoupled (softmax activation replaced by sigmoid). This can impose difficulty in establishing a fixed global threshold for computing F1-scores. In contrast, mAP scores describe the area under the precision-recall curve and do not depend on a fixed threshold. They provide a more holistic measure to summarize the class-wise predictive performance of a model by considering the relative ordering of per-class probabilities only. In Fig. 6, we demonstrate a scenario for which F1-scores fail to describe the actual predictive capacity. While for *additive noise*, the scores of mAP-micro and F1-micro correlate, any noise scenario including *subtractive noise* (pure or mixed) causes a large drop in F1-score performance, including the datasets of UCMerced and BEN19-Ireland, for which mAP-micro scores appear to be more stable. The presence of *subtractive noise* in MLC prevents the probabilities of underrepresented classes from scaling well between 0 and 1; instead, they stay close to 0. Still, reasonable decision boundaries are maintained at smaller thresholds with the optimal value differing between classes. A metric like F1-scores relying on a fixed global threshold fails to summarize the predictive capacity of a method under these class-specific artifacts. Therefore, we abandon the F1-score metric for further experiments. Instead, we report the mAP metric only.

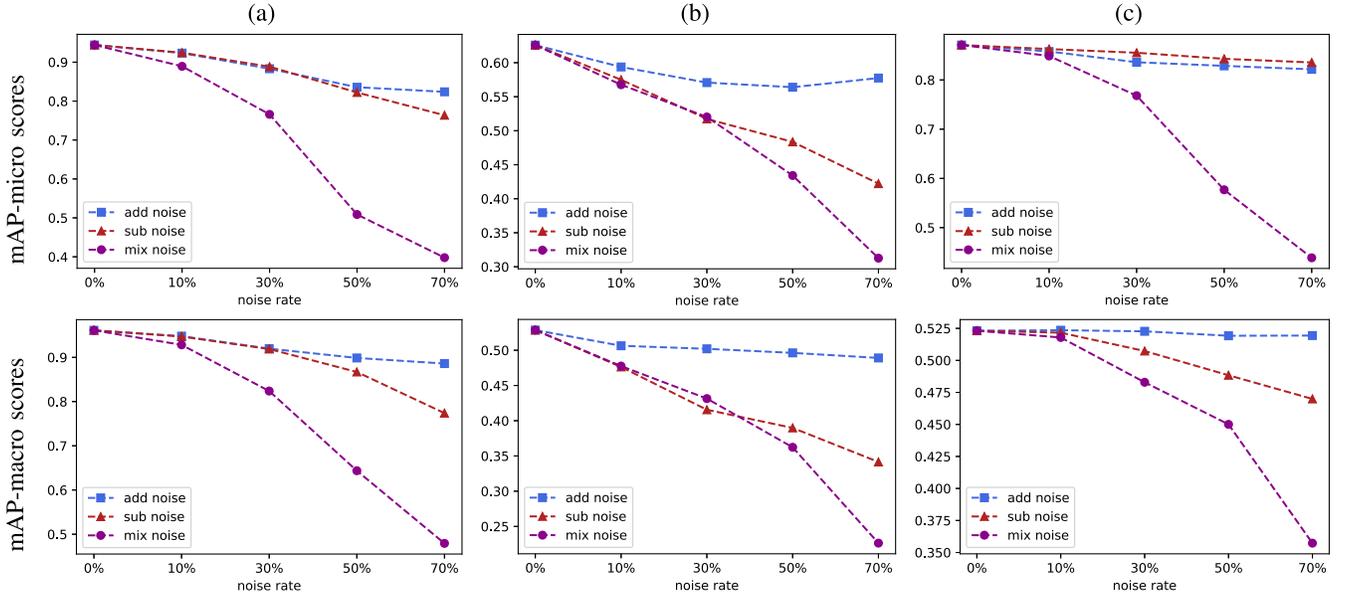

Fig. 5. Effects of different noise types under different noise rates compared for three multi-label RS datasets: (a) UCMerced; (b) TreeSatAI; and (c) BEN19-Ireland. Backbone ResNet18 with binary cross entropy. *Additive noise* denoted as add noise, *subtractive noise* denoted as sub noise and *mixed noise* denoted as mix noise.

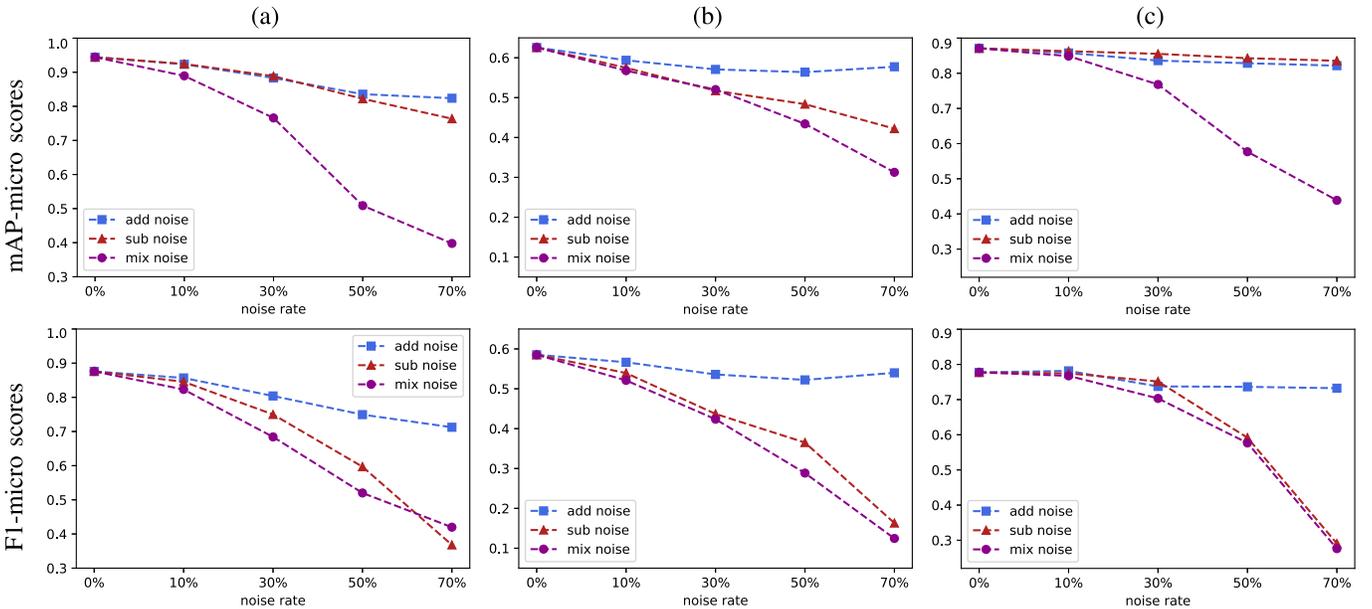

Fig. 6. Comparison of metrics mAP-micro and F1-micro under different noise rates compared for three different multi-label RS datasets: (a) UCMerced; (b) TreeSatAI; and (c) BEN19-Ireland. Backbone ResNet18 with Binary Cross Entropy. *Additive noise* denoted as add noise, *subtractive noise* denoted as sub noise and *mixed noise* denoted as mix noise.

### D. Comparison of Adapted Noise Robust Methods

Tables III–V show the best mAP-micro and mAP-macro scores for three datasets under different noise rates evaluated for the baseline model as well as the adapted noise robust methods. In general, it can be noticed that all adapted methods

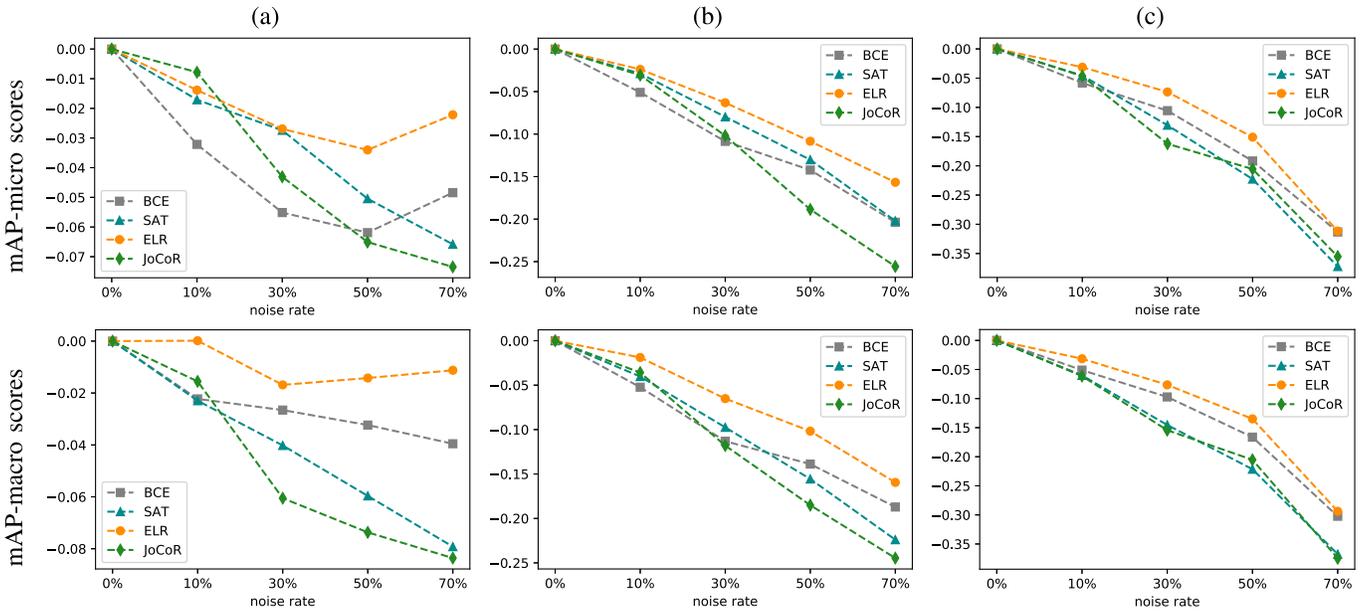

Fig. 7. Performance drop with respect to 0% noise of adapted noise robust methods for different noise types under different noise rates for the TreeSatAI dataset: (a) *additive noise*; (b) *subtractive noise*; and (c) *mixed noise*.

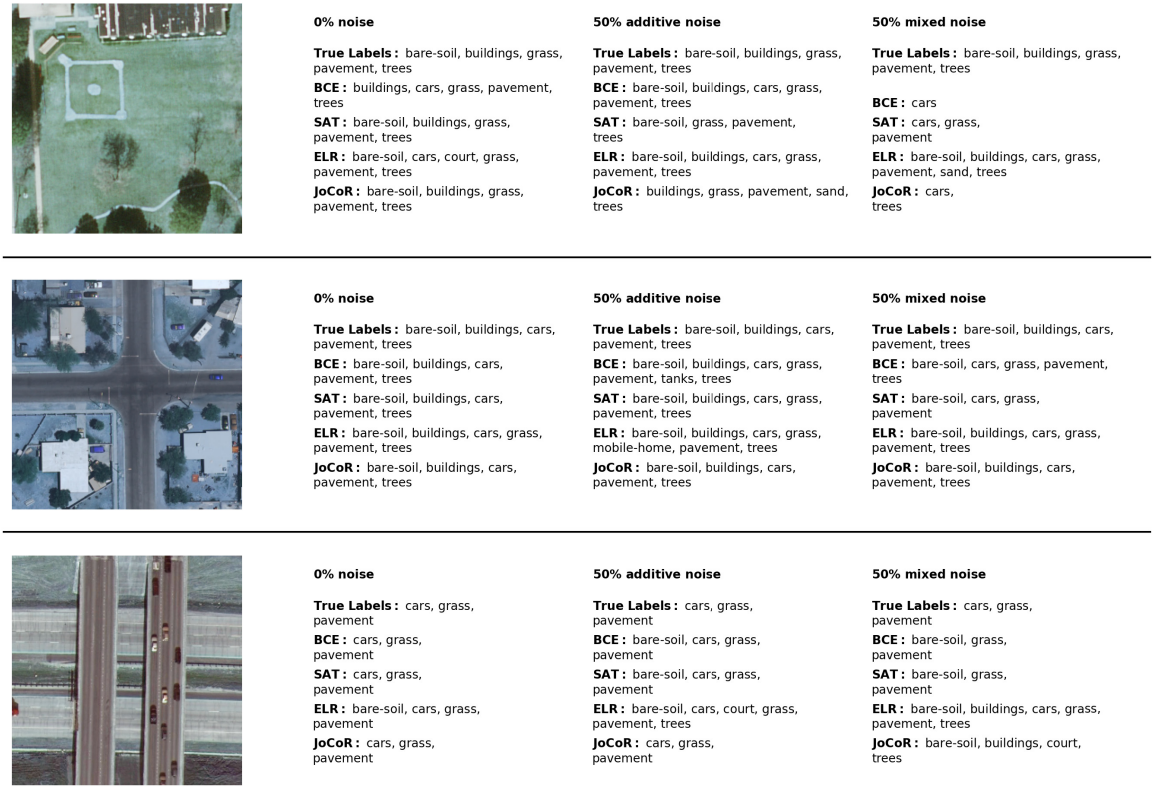

Fig. 8. Qualitative results of predictions on three randomly sampled images from the test set of UCMerced dataset. *True label* relates to the clean label of the images. Columns indicate the noise rate injected to the training set on which the methods were trained on.

reveal almost analogous behavior to different types and rates of noise, following baseline similar functions of performance decrease for higher injection rates of synthetic noise. For the BEN19-Ireland dataset (see Table V), none of the adapted methods produces significantly better results than the baseline. In fact, the scores are alike for all methods at all noise rates and for all noise types. For the other two datasets, the adapted methods show more noise robustness compared to the baseline.

TABLE III

RESULTS IN MAP MICRO/MACRO UNDER (a) *Additive Noise*; (b) *Subtractive Noise*; AND (c) *Mixed Noise* FOR THE UCMERCED DATASET

(a)

| Method | Additive Noise | | | | |
|---|---|---|---|---|---|
| | 0% | 10% | 30% | 50% | 70% |
| BCE | 0.94/0.96 | 0.92/0.95 | 0.88/0.92 | 0.84/0.90 | 0.82/0.89 |
| SAT | 0.96/0.97 | 0.95/0.96 | 0.93/0.95 | 0.88/0.93 | 0.84/0.92 |
| ELR | 0.95/0.96 | 0.93/0.95 | 0.90/0.93 | 0.87/0.92 | 0.84/0.90 |
| JoCoR | 0.94/0.96 | 0.93/0.95 | 0.88/0.92 | 0.85/0.92 | 0.81/0.88 |

(b)

| Method | Subtractive Noise | | | | |
|---|---|---|---|---|---|
| | 0% | 10% | 30% | 50% | 70% |
| BCE | 0.94/0.96 | 0.92/0.95 | 0.89/0.92 | 0.82/0.87 | 0.76/0.77 |
| SAT | 0.96/0.97 | 0.94/0.96 | 0.92/0.94 | 0.89/0.91 | 0.83/0.85 |
| ELR | 0.95/0.96 | 0.92/0.95 | 0.91/0.93 | 0.87/0.90 | 0.80/0.82 |
| JoCoR | 0.94/0.96 | 0.93/0.95 | 0.90/0.92 | 0.82/0.86 | 0.75/0.76 |

(c)

| Method | Mixed Noise | | | | |
|---|---|---|---|---|---|
| | 0% | 10% | 30% | 50% | 70% |
| BCE | 0.94/0.96 | 0.89/0.93 | 0.77/0.82 | 0.51/0.64 | 0.40/0.48 |
| SAT | 0.96/0.97 | 0.92/0.94 | 0.79/0.88 | 0.51/0.68 | 0.37/0.52 |
| ELR | 0.95/0.96 | 0.91/0.94 | 0.77/0.87 | 0.52/0.70 | 0.37/0.52 |
| JoCoR | 0.94/0.96 | 0.92/0.95 | 0.75/0.82 | 0.51/0.66 | 0.38/0.48 |

TABLE IV

RESULTS IN MAP MICRO/MACRO UNDER (a) *Additive Noise*; (b) *Subtractive Noise*; AND (c) *Mixed Noise* FOR THE TREESATAI DATASET

(a)

| Method | Additive Noise | | | | |
|---|---|---|---|---|---|
| | 0% | 10% | 30% | 50% | 70% |
| BCE | 0.63/0.53 | 0.59/0.51 | 0.57/0.50 | 0.56/0.50 | 0.58/0.49 |
| SAT | 0.69/0.60 | 0.67/0.57 | 0.66/0.56 | 0.64/0.54 | 0.63/0.52 |
| ELR | 0.63/0.53 | 0.62/0.53 | 0.60/0.51 | 0.60/0.52 | 0.61/0.52 |
| JoCoR | 0.67/0.59 | 0.66/0.57 | 0.62/0.53 | 0.60/0.52 | 0.59/0.51 |

(b)

| Method | Subtractive Noise | | | | |
|---|---|---|---|---|---|
| | 0% | 10% | 30% | 50% | 70% |
| BCE | 0.63/0.53 | 0.57/0.48 | 0.52/0.42 | 0.48/0.39 | 0.42/0.34 |
| SAT | 0.69/0.60 | 0.66/0.56 | 0.61/0.50 | 0.56/0.44 | 0.49/0.37 |
| ELR | 0.63/0.53 | 0.61/0.51 | 0.57/0.47 | 0.52/0.43 | 0.47/0.37 |
| JoCoR | 0.67/0.59 | 0.64/0.55 | 0.57/0.47 | 0.48/0.40 | 0.41/0.34 |

(c)

| Method | Mixed Noise | | | | |
|---|---|---|---|---|---|
| | 0% | 10% | 30% | 50% | 70% |
| BCE | 0.63/0.53 | 0.57/0.48 | 0.52/0.43 | 0.43/0.36 | 0.31/0.23 |
| SAT | 0.69/0.60 | 0.65/0.54 | 0.56/0.45 | 0.47/0.38 | 0.32/0.23 |
| ELR | 0.63/0.53 | 0.60/0.50 | 0.56/0.45 | 0.48/0.40 | 0.32/0.24 |
| JoCoR | 0.67/0.59 | 0.62/0.53 | 0.50/0.43 | 0.46/0.38 | 0.31/0.21 |

TABLE V

RESULTS IN MAP MICRO/MACRO UNDER (a) *Additive Noise*; (b) *Subtractive Noise*; AND (c) *Mixed Noise* FOR THE BEN19-IRELAND DATASET

(a)

| Method | Additive Noise | | | | |
|---|---|---|---|---|---|
| | 0% | 10% | 30% | 50% | 70% |
| BCE | 0.87/0.52 | 0.86/0.52 | 0.84/0.52 | 0.83/0.52 | 0.82/0.52 |
| SAT | 0.87/0.54 | 0.86/0.54 | 0.84/0.52 | 0.83/0.52 | 0.82/0.52 |
| ELR | 0.87/0.52 | 0.86/0.53 | 0.84/0.54 | 0.83/0.53 | 0.82/0.53 |
| JoCoR | 0.87/0.51 | 0.87/0.52 | 0.84/0.51 | 0.83/0.52 | 0.82/0.52 |

(b)

| Method | Subtractive Noise | | | | |
|---|---|---|---|---|---|
| | 0% | 10% | 30% | 50% | 70% |
| BCE | 0.87/0.52 | 0.86/0.52 | 0.86/0.51 | 0.84/0.49 | 0.84/0.47 |
| SAT | 0.87/0.54 | 0.86/0.53 | 0.85/0.51 | 0.85/0.49 | 0.83/0.46 |
| ELR | 0.87/0.52 | 0.87/0.52 | 0.86/0.52 | 0.86/0.50 | 0.85/0.47 |
| JoCoR | 0.87/0.51 | 0.86/0.52 | 0.85/0.51 | 0.85/0.48 | 0.84/0.46 |

(c)

| Method | Mixed Noise | | | | |
|---|---|---|---|---|---|
| | 0% | 10% | 30% | 50% | 70% |
| BCE | 0.87/0.52 | 0.85/0.52 | 0.77/0.48 | 0.58/0.45 | 0.44/0.36 |
| SAT | 0.87/0.54 | 0.85/0.52 | 0.76/0.48 | 0.58/0.44 | 0.43/0.36 |
| ELR | 0.87/0.52 | 0.85/0.52 | 0.73/0.49 | 0.58/0.46 | 0.43/0.37 |
| JoCoR | 0.87/0.51 | 0.85/0.51 | 0.77/0.48 | 0.58/0.45 | 0.44/0.36 |

In particular, the adapted versions of ELR and SAT outperform the baseline and JoCoR for the UCMerced dataset for all types of noise, most notably under high *subtractive noise* (see Table III). A similar trend can be noticed when looking at the results of the TreeSatAI dataset (see Table IV). Here, also the JoCoR method seems to be more noise robust than the baseline. However, since the initial scores at 0% artificially injected noise differ by more than 0.05 points, looking at the final performance only is not sufficient to evaluate noise robustness. Fig. 7 adopts a different perspective on noise robustness by visualizing the differences for each noise rate with respect to the initial performance at 0% noise for the TreeSatAI dataset. In particular, ELR can be identified as the most robust method to multi-label noise. Likewise, for the UCMerced dataset, the difference values indicate superior noise robustness for ELR, while for the BEN19-Ireland dataset, none of the methods consistently shows such superiority. Based on all obtained results on three datasets, we observe that the adapted SAT is able to reach the highest scores under label noise, while the adapted ELR method reveals the strongest robustness to multi-label noise. A qualitative analysis of predictions on the clean test set of the UCMerced dataset further emphasizes our findings (see Fig. 8). Under 0% noise and 50% *additive noise* (columns 2 and 3), the predictions of all methods are mostly in line with the clean ground reference samples. However, under the more challenging setup of 50% *mixed noise* (column 4), whose increased difficulty can be attributed to the *subtractive noise* present in the training set, ELR is more reliable in predicting the present classes. In this scenario, the other methods tend to generally make less and rather incorrect predictions. Nonetheless, it has to be noted that the drawback of more correctly predicted present classes of ELR is the additional false prediction of classes that are not present in the image (e.g., 50% *mixed noise*, bottom row: bare-soil and trees). Interestingly, ELR also seems to make more confident decisions in ambiguous situations without any artificial noise injection. Even though the clean class label of the bottom image does not indicate bare-soil, a visual examination of the image does not fully exclude the possibility that the lower part of the image actually covers some bare-soil. In opposition to the clean labels, the ELR is the only method that predicts this ambiguous class as present in the image. In general, given the higher complexity of predicting multiple labels in comparison with SLC tasks, we think that the higher noise robustness of the ELR method can be due to a higher fine-grain control on learning signals from potentially noisy labels of a sample. Instead of entirely discarding individual labels (JoCoR) or smoothing all labels (SAT), the ELR regularization operates on the continuous spectrum, only slowly increasing the regularization term of the labels suspect to contain noise (and hence decreasing their loss). An erroneously regularized clean label may still be able to be recovered during a later stage of the learning process. Above all, this may be particularly relevant for datasets resembling high label imbalances, in which the dominant classes are learned faster.

## VI. CONCLUSION

In this article, we have adapted three suitable single-label noise robust methods from CV for handling multi-label noise in RS. We based our selection process on a discussion of the potential and limitations of single-label methods in multi-label noise scenarios. We have adapted one sample selection method that derives its decisions from multi-network learning (JoCoR) and two loss adjustment methods that robustify learning via label refurbishment (SAT) and via a regularization term (ELR). We have evaluated the adapted methods on three RS multi-label datasets: 1) UCMerced; 2) TreeSatAI; and 3) BEN19-Ireland. Unlike the existing works, we have proposed to study different types of multi-label noise (*additive* and *subtractive*) separately and jointly (*mixed*) with a balanced contribution of both noise types. In general, we have shown that RS training sets with a low fraction of present labels containing any form of *subtractive noise* (pure or mixed) have a significantly stronger negative effect on the performance of deep neural networks than purely *additive noise*. RS training sets with a higher fraction of present labels only become difficult to learn from when *additive noise* and *subtractive noise* are present at the same time. The results of the adapted methods indicate a stronger boost to performance and robustness when refurbishing labels (SAT) or adding a label-specific regularization term (ELR), counteracting label noise. Except for pure noise for the TreeSatAI dataset, label selection based multi-network learning could not improve performance with respect to our baseline. While pure *additive noise* and pure *subtractive noise* have less negative effect on model performance in general (baseline and adapted methods), the mutual occurrence still remains an open challenge. Not only does *mixed noise* decrease the learning performance disproportionately, but it is also the noise type to which the adapted methods showed the least additional robustness. For practical scenarios, the results indicate a preferable use of the SAT method when a small amount of label noise is present, while in extreme label noise regimes, the application of the intrinsically more noise robust ELR method should be favored. Further, by comparing metric results under *subtractive noise*, we have illustrated a scenario for which F1-score is not a sufficient metric to summarize model performance on multi-labeled data. As a future work, we plan on developing a method robust to multi-label noise that emphasizes robustness toward *mixed noise*. As a final remark, it is recommended to favor the occurrence of *additive noise* over *subtractive noise* when creating multi-label datasets under the risk of label noise.

## REFERENCES

[1] G. Sumbul and B. Demir, "A deep multi-attention driven approach for multi-label remote sensing image classification," *IEEE Access*, vol. 8, pp. 95934–95946, 2020.

[2] P. Sobti, A. Nayyar, Niharika, and P. Nagrath, "EnsemV3X: A novel ensembled deep learning architecture for multi-label scene classification," *PeerJ Comput. Sci.*, vol. 7, p. e557, May 2021.

[3] Y. Li, R. Chen, Y. Zhang, M. Zhang, and L. Chen, "Multi-label remote sensing image scene classification by combining a convolutional neural network and a graph neural network," *Remote Sensing*, vol. 23, p. 4003, Dec. 2020.

[4] G. Büttner, J. Feranec, G. Jaffrain, L. Mari, G. Maucha, and T. Soukup, "The CORINE land cover 2000 project," *EARSeL eProceedings*, vol. 3, no. 3, pp. 331–346, 2004.

[5] E. Bartholomé and A. S. Belward, "GLC2000: A new approach to global land cover mapping from Earth observation data," *Int. J. Remote Sens.*, vol. 26, no. 9, pp. 1959–1977, 2005.

[6] O. Arino, J. J. R. Perez, V. Kalogirou, S. Bontemps, P. Defourny, and E. V. Bogaert, "Global land cover map for 2009 (GlobCover 2009)," European Space Agency (ESA), Université Catholique de Louvain (UCL), 2012.

[7] L. Jian, F. Gao, P. Ren, Y. Song, and S. Luo, "A noise-resilient online learning algorithm for scene classification," *Remote Sens.*, vol. 10, no. 11, p. 1836, Nov. 2018.

[8] B. B. Damodaran, R. Flamary, V. Seguy, and N. Courty, "An entropic optimal transport loss for learning deep neural networks under label noise in remote sensing images," *Comput. Vis. Image Understand.*, vol. 191, Feb. 2020, Art. no. 102863.

[9] Y. Li, Y. Zhang, and Z. Zhu, "Error-tolerant deep learning for remote sensing image scene classification," *IEEE Trans. Cybern.*, vol. 51, no. 4, pp. 1756–1768, Apr. 2020.

[10] Y. Hua, S. Lobry, L. Mou, D. Tuia, and X. X. Zhu, "Learning multi-label aerial image classification under label noise: A regularization approach using word embeddings," in *Proc. IEEE Int. Geosci. Remote Sens. Symp. (IGARSS)*, Sep. 2020, pp. 525–528.

[11] A. K. Aksoy, M. Ravanbakhsh, and B. Demir, "Multi-label noise robust collaborative learning for remote sensing image classification," *IEEE Trans. Neural Netw. Learn. Syst.*, early access, Oct. 20, 2022, doi: 10.1109/TNNLS.2022.3209992.

[12] L. Huang, C. Zhang, and H. Zhang, "Self-adaptive training: Beyond empirical risk minimization," in *Proc. Adv. Neural Inf. Process. Syst.*, vol. 33. Red Hook, NY, USA: Curran Associates, 2020, pp. 19365–19376.

[13] S. Liu, J. Niles-Weed, N. Razavian, and C. Fernandez-Granda, "Early-learning regularization prevents memorization of noisy labels," in *Proc. NeurIPS*, vol. 33. Red Hook, NY, USA: Curran Associates, 2020, pp. 20331–20342.

[14] H. Wei, L. Feng, X. Chen, and B. An, "Combating noisy labels by agreement: A joint training method with co-regularization," in *Proc. IEEE/CVF Conf. Comput. Vis. Pattern Recognit. (CVPR)*, Jun. 2020, pp. 13723–13732.

[15] D. Hendrycks, M. Mazeika, D. Wilson, and K. Gimpel, "Using trusted data to train deep networks on labels corrupted by severe noise," in *Advances in Neural Information Processing Systems*, vol. 31. Red Hook, NY, USA: Curran Associates, 2018.

[16] K.-H. Lee, X. He, L. Zhang, and L. Yang, "CleanNet: Transfer learning for scalable image classifier training with label noise," in *Proc. IEEE/CVF Conf. Comput. Vis. Pattern Recognit.*, Jun. 2018, pp. 5447–5456.

[17] Y. Li, J. Yang, Y. Song, L. Cao, J. Luo, and L.-J. Li, "Learning from noisy labels with distillation," in *Proc. IEEE Int. Conf. Comput. Vis. (ICCV)*, Oct. 2017, pp. 1928–1936.

[18] G. Song and W. Chai, "Collaborative learning for deep neural networks," in *Proc. Adv. Neural Inf. Process. Syst.*, vol. 31. Red Hook, NY, USA: Curran Associates, 2018, pp. 1832–1841.

[19] S. Sukhbaatar, J. Bruna, M. Paluri, L. Bourdev, and R. Fergus, "Training convolutional networks with noisy labels," 2014, *arXiv:1406.2080*.

[20] A. Ghosh, H. Kumar, and P. Sastry, "Robust loss functions under label noise for deep neural networks," in *Proc. AAAI Conf. Artif. Intell.*, vol. 31, no. 1, 2017, pp. 1919–1925.

[21] Z. Zhang and M. Sabuncu, "Generalized cross entropy loss for training deep neural networks with noisy labels," in *Advances in Neural Information Processing Systems*, vol. 31. Red Hook, NY, USA: Curran Associates, 2018.

[22] H. Zhang, M. Cisse, Y. N. Dauphin, and D. Lopez-Paz, "Mixup: Beyond empirical risk minimization," in *Proc. Int. Conf. Learn. Represent.*, 2018, pp. 1–13.

[23] S. Guo et al., "Curriculumnet: Weakly supervised learning from large-scale web images," in *Proc. Eur. Conf. Comput. Vis.*, 2018, pp. 135–150.

[24] D. T. Nguyen, C. K. Mummadi, T. P. N. Ngo, T. H. P. Nguyen, L. Beggel, and T. Brox, "SELF: Learning to filter noisy labels with self-ensembling," in *Proc. Int. Conf. Learn. Represent.*, 2020, pp. 1–15.

[25] E. Malach and S. Shalev-Shwartz, "Decoupling 'when to update' from 'how to update,'" in *Advances in Neural Information Processing Systems*, vol. 30. Red Hook, NY, USA: Curran Associates, 2017.

[26] Y. Han, S. K. Roy, L. Petersson, and M. Harandi, "Learning from noisy labels via discrepant collaborative training," in *Proc. IEEE Winter Conf. Appl. Comput. Vis. (WACV)*, Mar. 2020, pp. 3158–3167.

[27] C. G. Northcutt, T. Wu, and I. L. Chuang, "Learning with confident examples: Rank pruning for robust classification with noisy labels," 2017, *arXiv:1705.01936*.

[28] B. Han et al., "Co-teaching: Robust training of deep neural networks with extremely noisy labels," in *Advances in Neural Information Processing Systems*, vol. 31. Red Hook, NY, USA: Curran Associates, 2018.

[29] L. Jiang, Z. Zhou, T. Leung, L.-J. Li, and L. Fei-Fei, "MentorNet: Learning data-driven curriculum for very deep neural networks on corrupted labels," in *Proc. Int. Conf. Mach. Learn.*, 2018, pp. 2304–2313.

[30] J. Han, P. Luo, and X. Wang, "Deep self-learning from noisy labels," in *Proc. IEEE/CVF Int. Conf. Comput. Vis. (ICCV)*, Oct. 2019, pp. 5137–5146.

[31] J. Li, Y. Wong, Q. Zhao, and M. S. Kankanhalli, "Learning to learn from noisy labeled data," in *Proc. IEEE/CVF Conf. Comput. Vis. Pattern Recognit. (CVPR)*, Jun. 2019, pp. 5051–5059.

[32] H. Song, M. Kim, D. Park, Y. Shin, and J.-G. Lee, "Learning from noisy labels with deep neural networks: A survey," *IEEE Trans. Neural Netw. Learn. Syst.*, early access, Mar. 7, 2022, doi: 10.1109/TNNLS.2022.3152527.

[33] T. Durand, N. Mehrasa, and G. Mori, "Learning a deep ConvNet for multi-label classification with partial labels," in *Proc. IEEE/CVF Conf. Comput. Vis. Pattern Recognit. (CVPR)*, Jun. 2019, pp. 647–657.

[34] S. S. Bucak, R. Jin, and A. K. Jain, "Multi-label learning with incomplete class assignments," in *Proc. CVPR*, Jun. 2011, pp. 2801–2808.

[35] H. Jain, Y. Prabhu, and M. Varma, "Extreme multi-label loss functions for recommendation, tagging, ranking & other missing label applications," in *Proc. 22nd ACM SIGKDD Int. Conf. Knowl. Discovery Data Mining*, Aug. 2016, pp. 935–944.

[36] M.-K. Xie and S.-J. Huang, "Partial multi-label learning with noisy label identification," *IEEE Trans. Pattern Anal. Mach. Intell.*, vol. 44, no. 7, pp. 3676–3687, Jul. 2022.

[37] A. Veit, N. Alldrin, G. Chechik, I. Krasin, A. Gupta, and S. Belongie, "Learning from noisy large-scale datasets with minimal supervision," in *Proc. IEEE Conf. Comput. Vis. Pattern Recognit. (CVPR)*, Jul. 2017, pp. 839–847.

[38] M. Hu, H. Han, S. Shan, and X. Chen, "Multi-label learning from noisy labels with non-linear feature transformation," in *Proc. Asian Conf. Comput. Vis.* (Lecture Notes in Computer Science), vol. 11365. Cham, Switzerland: Springer, 2019, pp. 404–419, doi: 10.1007/978-3-030-20873-8_26.

[39] N. Inoue, E. Simo-Serra, T. Yamasaki, and H. Ishikawa, "Multi-label fashion image classification with minimal human supervision," in *Proc. IEEE Int. Conf. Comput. Vis. Workshops (ICCVW)*, Oct. 2017, pp. 2261–2267.

[40] H. Kumar, N. Manwani, and P. S. Sastry, "Robust learning of multi-label classifiers under label noise," in *Proc. 7th ACM IKDD CoDS 25th COMAD*, Jan. 2020, pp. 90–97.

[41] W. Zhao and C. Gomes, "Evaluating multi-label classifiers with noisy labels," 2021, *arXiv:2102.08427*.

[42] Z. Zhang, H. Zhang, S. O. Arik, H. Lee, and T. Pfister, "Distilling effective supervision from severe label noise," in *Proc. IEEE/CVF Conf. Comput. Vis. Pattern Recognit. (CVPR)*, Jun. 2020, pp. 9291–9300.

[43] J. Li, R. Socher, and S. C. Hoi, "DivideMix: Learning with noisy labels as semi-supervised learning," in *Proc. Int. Conf. Learn. Represent.*, 2019, pp. 1–14.

[44] K. He, X. Zhang, S. Ren, and J. Sun, "Deep residual learning for image recognition," in *Proc. IEEE Conf. Comput. Vis. Pattern Recognit. (CVPR)*, Jun. 2016, pp. 770–778.

[45] B. Chaudhuri, B. Demir, S. Chaudhuri, and L. Bruzzone, "Multilabel remote sensing image retrieval using a semisupervised graph-theoretic method," *IEEE Trans. Geosci. Remote Sens.*, vol. 56, no. 2, pp. 1144–1158, Feb. 2017.

[46] S. Ahlswede, N. T. Madam, C. Schulz, B. Kleinschmit, and B. Demir, "Weakly supervised semantic segmentation of remote sensing images for tree species classification based on explanation methods," in *Proc. IEEE Int. Geosci. Remote Sens. Symp. (IGARSS)*, Jul. 2022, pp. 4847–4850.

[47] G. Sumbul, M. Charfuelan, B. Demir, and V. Markl, "Bigearthnet: A large-scale benchmark archive for remote sensing image understanding," in *Proc. IEEE Int. Geosci. Remote Sens. Symp. (IGARSS)*, Jul. 2019, pp. 5901–5904.

[48] Y. Yang and S. Newsam, "Bag-of-visual-words and spatial extensions for land-use classification," in *Proc. 18th SIGSPATIAL Int. Conf. Adv. Geographic Inf. Syst. (GIS)*, 2010, pp. 270–279.

[49] G. Sumbul et al., "BigEarthNet-MM: A large-scale, multimodal, multilabel benchmark archive for remote sensing image classification and retrieval [software and data sets]," *IEEE Geosci. Remote Sens. Mag.*, vol. 9, no. 3, pp. 174–180, Sep. 2021.

[50] J. Deng, W. Dong, R. Socher, L.-J. Li, K. Li, and L. Fei-Fei, "ImageNet: A large-scale hierarchical image database," in *Proc. IEEE Conf. Comput. Vis. Pattern Recognit.*, Jun. 2009, pp. 248–255.

[51] I. Loshchilov and F. Hutter, "Decoupled weight decay regularization," 2017, *arXiv:1711.05101*.

[52] M. Everingham, L. Van Gool, C. K. I. Williams, J. Winn, and W. Zisserman, "The PASCAL visual object classes (VOC) challenge," *Int. J. Comput. Vis.*, vol. 88, no. 2, pp. 303–338, Sep. 2010.

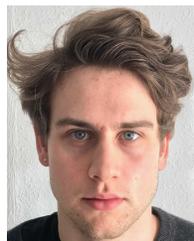

**Tom Burgert** (Member, IEEE) received the M.Sc. degree in computer science from Technische Universität (TU) Berlin, Berlin, Germany, in 2022, where he is currently pursuing the Ph.D. degree in machine learning with the Remote Sensing and Image Analysis (RSiM) Group.

He joined the RSiM Group as a Student Research Assistant during his master's study. His research interests evolve around learning theory and explaining deep neural networks in the intersection of computer vision and remote sensing.

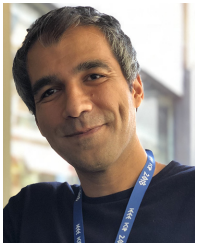

**Mahdyar Ravanbakhsh** (Member, IEEE) received the Ph.D. degree from the University of Genoa, Genoa, Italy, in 2019.

He was a Post-Doctoral Research Fellow with the University of Genoa, and a Visiting Researcher with the Deep Relational Learning Group, University of Trento, Trento, Italy, in 2016. He was a Research Fellow of the Remote Sensing and Image Analysis (RSiM) Group, Department of Electrical Engineering and Computer Science, Technische Universität (TU) Berlin, Berlin, Germany, from 2019 to 2022. He is a Scientific Researcher with Zalando SE, Berlin. His research interests include the intersection of machine learning and computer vision with an emphasis on deep learning with minimal supervision and/or limited data.

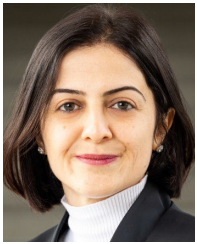

**Begüm Demir** (Senior Member, IEEE) received the B.S., M.Sc., and Ph.D. degrees in electronic and telecommunication engineering from Kocaeli University, Kocaeli, Turkey, in 2005, 2007, and 2010, respectively.

She is currently a Full Professor and the Founder Head of the Remote Sensing Image Analysis (RSiM) Group, Faculty of Electrical Engineering and Computer Science, Technische Universität (TU) Berlin, Berlin, Germany, and also the Head of the Big Data Analytics for Earth Observation Research Group, Berlin Institute for the Foundations of Learning and Data (BIFOLD), Berlin. Her research interests include the intersection of machine learning, remote sensing, and signal processing. Specifically, she performs research in the field of processing and analysis of large-scale earth observation data acquired by airborne and satellite-borne systems.

Dr. Demir is a Scientific Committee Member of several international conferences and workshops, such as Conference on Content-Based Multimedia Indexing, Conference on Big Data from Space, Living Planet Symposium, International Joint Urban Remote Sensing Event, International Society for Optics and Photonics (SPIE) International Conference on Signal and Image Processing for Remote Sensing, and Machine Learning for Earth Observation Workshop organized within the European Conference on Machine Learning and Principles and Practice of Knowledge Discovery in Databases (ECML PKDD). She is a Referee for several journals, such as PROCEEDINGS OF THE IEEE, IEEE TRANSACTIONS ON GEOSCIENCE AND REMOTE SENSING, IEEE GEOSCIENCE AND REMOTE SENSING LETTERS, IEEE TRANSACTIONS ON IMAGE PROCESSING, *Pattern Recognition*, IEEE TRANSACTIONS ON CIRCUITS AND SYSTEMS FOR VIDEO TECHNOLOGY, IEEE JOURNAL OF SELECTED TOPICS IN SIGNAL PROCESSING, *International Journal of Remote Sensing*, and several international conferences. She was awarded by the prestigious "2018 Early Career Award" by the IEEE Geoscience and Remote Sensing Society for her research contributions in machine learning for information retrieval in remote sensing. In 2018, she received a Starting Grant from the European Research Council (ERC) for her project "BigEarth: Accurate and Scalable Processing of Big Data in Earth Observation." She is a fellow of the European Laboratory for Learning and Intelligent Systems (ELLIS). She is currently an Associate Editor of IEEE GEOSCIENCE AND REMOTE SENSING LETTERS, *MDPI Remote Sensing*, and *International Journal of Remote Sensing*.